\documentclass[10pt,logo,copyright]{nvidia_release_template/nvidia_report}
\RequireXeTeX
\usepackage[numbers,sort&compress]{natbib}
\usepackage[noend]{algpseudocode}
\usepackage{algorithm}
\usepackage{amsmath}
\usepackage{amssymb}
\usepackage[normalem]{ulem}
\usepackage{wrapfig}
\usepackage{bm}
\usepackage{xcolor}
\usepackage{graphicx}
\usepackage{hyperref}
\usepackage{booktabs}

\newcommand{\proc}[1]{\textnormal{\texttt{#1}}}

\newcommand{\kw}[1]{{\bf #1}}

\newcommand{\R}{\mathbf{R}}

\newcommand{\SE}[1]{\mathrm{SE}(#1)}
\newcommand{\T}[2]{T^{#1}_{#2}}

\newcommand{\suchthat}{\quad\text{s.t.}\quad}
\newcommand{\decision}[1]{\mathbf{#1}}

\usepackage{amsthm}
\theoremstyle{plain}
\theoremstyle{definition}
\theoremstyle{plain}
\theoremstyle{plain}

\usepackage{listings}
\definecolor{deepblue}{rgb}{0,0,0.5}
\definecolor{deepred}{rgb}{0.6,0,0}
\definecolor{magenta}{rgb}{1.0,0,1.0}
\definecolor{deepgreen}{rgb}{0,0.5,0}
\definecolor{textblue}{rgb}{.2,.2,.7}
\definecolor{textred}{rgb}{0.54,0,0}
\definecolor{textgreen}{rgb}{0,0.43,0}
\definecolor{es-blue}{rgb}{0.1372,0.666,1}
\definecolor{lightgraybg}{RGB}{247,247,247}  %

\lstdefinestyle{pythonstyle}{
    language=Python, 
    breaklines=true,
    basicstyle=\ttfamily\small,
    emphstyle=\bfseries\color{deepred}, 
    emph={forward,forward_v},         
    numbers=left,
    numberstyle=\tiny\color{gray},
    stepnumber=1,
    numbersep=12pt,
    tabsize=2,
    stringstyle=\color{textgreen},
    frame=none,                    
    columns=fullflexible,
    keepspaces=true,
    xleftmargin=\parindent,
    showstringspaces=false,
    commentstyle=\color{deepgreen},
    keywordstyle=\color{es-blue},
    backgroundcolor=\color{lightgraybg}  %
}

\newcommand{\approach}{ScheduleGen}

\usepackage{placeins}
\usepackage{needspace}
\graphicspath{{figures/}}
\hypersetup{
 pdftitle={ScheduleGen: Demonstration Generation via Skill Scheduling for Multi-Arm Imitation Learning},
 pdfauthor={Neel Jawale; Stephan Pleines; Soha Pouya; Yuke Zhu; Ajay Mandlekar; Caelan Garrett},
 pdfsubject={Multi-arm imitation learning and robot demonstration generation},
 pdfkeywords={robot learning, imitation learning, multi-arm manipulation, data generation, scheduling},
 bookmarksnumbered=true
}
\title{ScheduleGen: Demonstration Generation via Skill Scheduling for Multi-Arm Imitation Learning}
\author{\parbox{\textwidth}{\centering
{\small\bfseries Neel Jawale$^{1}$, Stephan Pleines$^{1}$, Soha Pouya$^{1}$, Yuke Zhu$^{1,2}$, Ajay Mandlekar$^{1,\dagger}$, Caelan Garrett$^{1,\dagger}$}\\
{\footnotesize $^{1}$NVIDIA \quad $^{2}$The University of Texas at Austin \quad $^{\dagger}$Equal advising}
}}
\renewcommand{\keywordsfont}{\normalfont\itshape\fontsize{9}{11}\selectfont}
\begin{document}
\maketitle
\begin{figure}[!htbp]
\centering
\includegraphics[width=\linewidth]{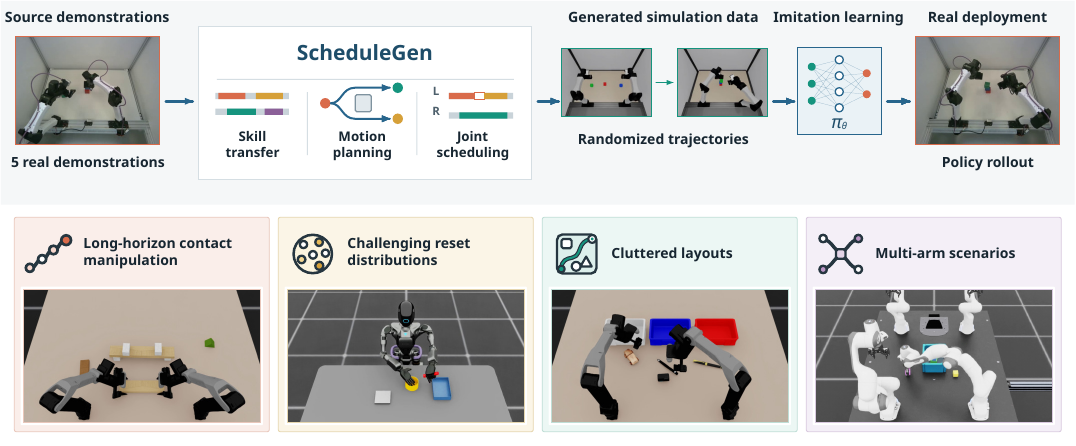}
\caption{\textbf{ScheduleGen overview.} {\it Top:} From a few source demonstrations, ScheduleGen adapts contact-rich skill demonstrations, plans collision-free connecting motions, and jointly schedules multiple arms to generate simulation data for multi-arm imitation learning that transfers to the real world. {\it Bottom:} ScheduleGen can handle long-horizon contact manipulation, challenging reset distributions, cluttered layouts, and multi-arm scenarios.}
\label{fig:overview}
\end{figure}
\begin{abstract}
Multi-arm robot platforms, including bimanual and humanoid robots, have the capability to complete tasks
with lower cycle times than single-arm systems, but they are more difficult to safely control due to the presence of multi-arm constraints.
The paradigm of training generalist robot policies from massive datasets is a promising approach to address general-purpose multi-arm manipulation, but the acquisition of large multi-arm manipulation datasets remains a bottleneck.
Automated demonstration generation can alleviate this burden by synthesizing new demonstrations from a handful of source demonstrations, but existing algorithms ignore multi-arm constraints and as a result often induce unsafe and unsuccessful behaviors.
We address this limitation by introducing ScheduleGen, an integrated planning \& scheduling algorithm for multi-arm demonstration generation.
We evaluate ScheduleGen on 28 simulation task variants, including seven externally authored RoboDojo tasks and settings inaccessible to prior systems; it achieves 79.6\% mean data-generation success and trains policies with 65.5\% mean success.
ScheduleGen policies outperform those trained on a state-of-the-art data generation baseline, and achieve 50--80\% success across eight real-world variants through zero-shot sim-to-real and real-to-sim-to-real transfer.
Videos and more at: \href{https://njawale42.github.io/ScheduleGen/}{ScheduleGen Website}
\end{abstract}

\begingroup
\abscontent
\par\endgroup
\normalfont\normalsize
\section{Introduction}

Generalist robot policies~\cite{black2024pi_0, bjorck2025gr00t} hold the promise of seamlessly solving a wide range of manipulation tasks across diverse robot embodiments. There is substantial interest in applying this paradigm to multi-arm embodiments, especially bimanual and humanoid robots, due to their ability to address a wider range of tasks compared to single-arm embodiments. Such embodiments make it possible to address tasks requiring multi-arm coordination and larger manipulation workspaces, and also have the potential to carry out tasks more efficiently due to simultaneous operation.

However, the predominant method for training generalist robot policies requires large-scale data collection, usually obtained through months of human teleoperation effort~\cite{ebert2021bridge, khazatsky2024droid}. Collecting such datasets is already painful and costly for single-arm embodiments, but these problems are exacerbated in the multi-arm setting, due to the need for special teleoperation interfaces~\cite{tung2021learning, zhao2023learning, darvish2023teleoperation}, and the resulting increase in operator burden. It can furthermore be difficult to capture desirable multi-arm coordination characteristics due to the limitations of the teleoperation interface~\cite{jiang2024dexmimicgen}.

Synthetic data generation in simulation is a compelling method to address this need. Prior methods allow for the generation of large-scale datasets with minimal human effort~\cite{mandlekar2023mimicgen, deshpande2026molmob0t} but fail to address some key challenges of generating data for multi-arm scenarios.
Some works use planning-based approaches~\cite{mcdonald2022guided, dalal2023imitating, deshpande2026molmob0t}, but contact-rich tasks remain difficult for such methods, and generating data for diverse multi-arm coordination characteristics remains challenging. DexMimicGen~\cite{jiang2024dexmimicgen} is able to handle different contact-rich tasks, but the coordination behaviors that it can generate are entirely dictated by the human demonstrations used to seed the process, limiting data diversity. Furthermore, the absence of planning means that it cannot handle situations where arm-arm or arm-object collisions are prevalent, such as wide object distributions or cluttered workspaces.

To address these shortcomings, we present ScheduleGen, a multi-arm manipulation data generation method that can encompass diverse multi-arm coordination strategies, long-horizon and contact-rich manipulation, and a wide range of workspaces and embodiments, including challenging object initializations and cluttered scenes. Given a small set of annotated demonstrations, \approach{} preserves object-relative contact skills, replans collision-free transitions, and jointly schedules the arms under task and collision constraints.

\textbf{The contributions of this paper are:}
\begin{itemize}
    \item ScheduleGen, a new automated demonstration generation algorithm for multi-arm systems that integrates planning, scheduling, and demonstration adaptation.
    \item ScheduleGen utilizes a novel planning formulation that first plans single-robot behavior to adapt object-centric skill demonstrations and then safely coordinates the robots by framing scheduling as a Mixed Integer Linear Program (MILP) with constraints evaluated using GPU-accelerated collision checking.
    \item We demonstrate ScheduleGen's applicability to long-horizon contact-rich tasks,  challenging resets, cluttered workspaces, and configurable coordination strategies across bimanual, four-arm, and humanoid systems. We evaluate ScheduleGen on 28 task variants across 3 simulation frameworks. On comparable tasks, \approach{}-trained policies outperform prior work DexMimicGen~\cite{jiang2024dexmimicgen}. In zero-shot sim-to-real and real-to-sim-to-real evaluation, the learned policies achieve 50--80\% success across eight real-world variants.

\end{itemize}

\begin{figure}[!t]
    \centering
    \includegraphics[width=\textwidth]{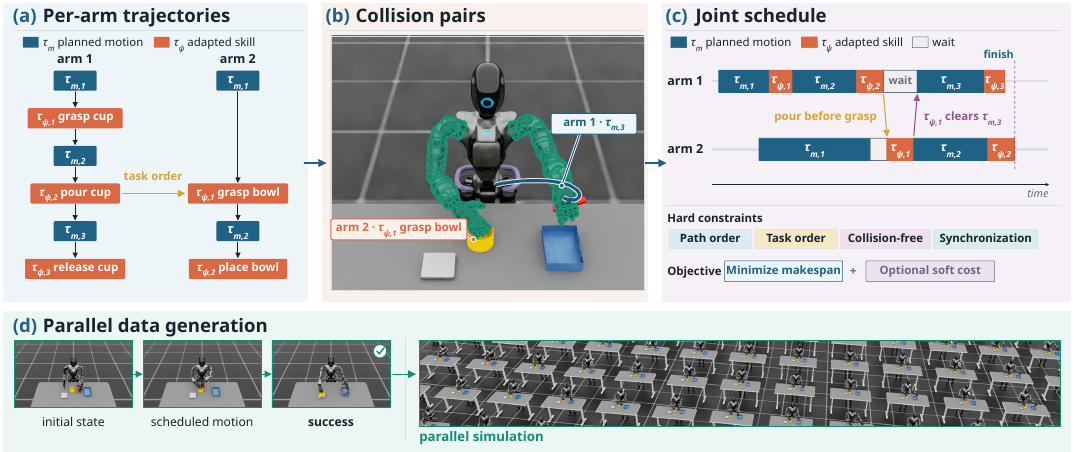}
    
    \caption{\textbf{ScheduleGen methodology.} (a) Contact-skill segments $\tau_\psi$ are adapted to the initial state and connected by planned motion $\tau_m$, producing partially ordered paths for each arm. (b) Collision checking identifies conflicting path segments. (c) ScheduleGen implicitly inserts waits to form a joint schedule that preserves task order, avoids collisions, and enforces synchronization when required. It minimizes makespan with optional soft costs, like summed robot execution times. (d) Scheduled motions are replayed across parallel simulation environments, and successful rollouts are retained as demonstrations.}
    \label{fig:methodology}
    
\end{figure}

\section{Related Work}

\textbf{Imitation Learning from Human Demonstrations.}
Manipulation demonstrations are often collected via robot teleoperation~\cite{zhang2017deep, mandlekar2018roboturk}, and
Behavioral Cloning (BC)~\cite{pomerleau1989alvinn}
is a common method for training manipulation policies on such datasets.
Recent efforts have scaled this paradigm to months of demonstration data using large teams of human teleoperators~\cite{ebert2021bridge, khazatsky2024droid}, and train expressive policy architectures on this data, resulting in capable generalist robot policies~\cite{black2024pi_0, bjorck2025gr00t}.
However, data sourcing remains a key bottleneck for this paradigm.
The problem is exacerbated for bimanual robots and humanoids, which can require specialized teleoperation interfaces to support multi-arm teleoperation~\cite{tung2021learning, zhao2023learning, darvish2023teleoperation}.

\textbf{Automated Demonstration Generation.}
Synthetic data generation in  simulation is a promising way to address the data bottleneck.
TAMP-based approaches~\cite{mcdonald2022guided,dalal2023imitating,garrett2021integrated} fully automate generation but are limited to single-robot prehensile manipulation.
MimicGen~\cite{mandlekar2023mimicgen} and DexMimicGen~\cite{jiang2024dexmimicgen} instead adapt a few teleoperated demonstrations to new initial states for single-arm and bimanual tasks, but adapting full demonstrations without accounting for collision or kinematic constraints limits their robustness to substantial initial-state variation.
SkillGen combines demonstration adaptation with motion planning to preserve contact-rich behavior while accommodating environmental and state changes~\cite{garrett2024skillgen}, but is limited to single-arm systems.
GLIDE~\cite{li2024planning} and HumanoidMimicGen~\cite{lin2026humanoidmimicgen} also incorporate motion planning and support some bimanual tasks, but independently planned arm motions compromise safety and success on complex bimanual tasks.
In contrast, ScheduleGen frames demonstration generation as a collision-aware scheduling problem to directly plan safe and efficient multi-arm manipulation.

\textbf{Multi-Arm Manipulation.}
Planning for multiple arms as a single composite robot can be intractable due to high-dimensional, combinatorial search spaces.
Multi-robot planners therefore factor the problem into interacting single-robot subproblems~\cite{dobson2017scalable}, reconciling pairwise constraints through backtracking over action timing~\cite{guo2024efficient}.
Prior work includes domain-specific planners for assembly~\cite{chen2022cooperative} and construction~\cite{hartmann2022long}, as well as single-robot TAMP frameworks extended to simultaneous multi-robot manipulation~\cite{Toussaint2017Multi-boundDomains,zhang2022mip,garrett2025schedulestream}.
ScheduleGen also uses factorization for efficient multi-arm planning, but it introduces demonstrations as schedulable skills and analyzes how scheduling decisions affect end-to-end visuomotor policy learning.

\section{Problem Description}\label{sec:problem}

Our objective is to develop an automated planning \& scheduling system that generates demonstration datasets for training multi-robot visuomotor policies using imitation learning.
We start by formulating our multi-robot control problem (Section~\ref{sec:prerequisites}).
Then, we define the notion of an object-centric single-robot manipulation skill (Section~\ref{sec:skills}).
Finally, we introduce temporal constraints among skills, which model the minimal set of temporal dynamics that skills must satisfy to correctly satisfy the task (Section~\ref{sec:constraints}).

\subsection{Prerequisites}\label{sec:prerequisites}

We are interested in controlling a set of robot arms ${\cal R}$ simultaneously to complete a task in an environment with objects represented by coordinate frames ${\cal F}$.
Let $w$ denote the world frame and $e_r$ denote robot $r$'s end-effector frame.
The state of each robot $r$ is represented by the {\em configuration} $q \in \R^d$ of its $d$ degrees of freedom (DOFs).
Let $\T{f_2}{f_1} \in \SE{3}$ denote the rigid transform from frame $f_1$ to frame $f_2$; thus, $\T{w}{f}$ is the {\em world pose} of frame $f$.
Let the pose of robot $r$'s end-effector frame $e_r$ be $\T{w}{e_r}$.

We model multi-robot manipulation as a Partially Observable Markov Decision Process (POMDP) with state space ${\cal S}$, action space ${\cal A}$, observation space ${\cal O}$, initial states $s_0 \sim p_0$, and goal states $S_* \subseteq {\cal S}$.
Each state $s$ encodes the configuration of each robot arm $s[r] = q$, pose of each end effector $s[e_r] = \T{w}{e_r}$, and pose of each object frame $s[f] = \T{w}{f}$.
Actions specify arm and finger targets.
Here, we use joint targets, but ScheduleGen also supports relative and absolute end-effector pose targets (Fig.~\ref{fig:methodology}).
Actions are executed by a composite controller at a constant rate of $1/\Delta t$.
Observations include external and wrist RGB camera images as well as centralized robot proprioception.
We wish to learn policies $\pi: {\cal O} \to {\cal A}$ that map observations to actions.

\subsection{Skill Adaptation}\label{sec:skills}

Our high-level demonstration generation strategy is to consume a set of object-centric skills per robot, adapt the skills given the current state of the objects, and schedule safe and correct execution of the skills among the robots.
Unlike prior work~\cite{garrett2024skillgen,lin2026humanoidmimicgen}, we support different methods for acquiring skills including 1) segmenting teleoperated demonstrations, 2) scripting waypoints, 3) using pre-grasp and grasp waypoints, and 4) agentically authoring skills.
A robot arm {\em skill} $\psi = \langle r, f, \vec{s}, \vec{a} \rangle$ is defined by a robot $r \in {\cal R}$, reference object frame $f \in {\cal F}$, sequence of states $\vec{s} \in {\cal S}^\infty$, and sequence of actions $\vec{a} \in {\cal A}^\infty$.

As a running example, consider the humanoid task in Figure~\ref{fig:methodology} where a humanoid with two robot arms is tasked with pouring from a red cup into a yellow bowl and then placing the bowl on a plate.
This example has five skills: three for arm 1 (grasp, pour, and release cup) and two for arm 2 (grasp and place bowl).
For this task, we acquired these skills by collecting bimanual demonstrations and then annotating intervals that correspond to each skill.

We adopt the procedure in MimicGen~\cite{mandlekar2023mimicgen} for adapting skills to new states by assuming approximate pose invariance.
Given a new state $s_0$, we adapt each task-space end-effector state $s[e_r]$ and action $a[e_r]$ relative to object frame $f$ using:
\[
s'[e_r] {\gets} (s_0[f] s[f]^{-1}) s[e_r],\qquad
a'[e_r] {\gets} (s_0[f] s[f]^{-1}) a[e_r].
\]
Section~\ref{sec:planning} converts adapted poses to joint targets through inverse kinematics or replays them directly in task space.
Adaptation does not modify finger states or actions. Appendix~\ref{app:path-construction} explains how adapted states and actions are used during planning and replay.

\subsection{Temporal Constraints}\label{sec:constraints}

In order to complete manipulation tasks, the team of robots will generally need to execute multiple skills.
When there's only a single robot, skills can only be executed sequentially~\cite{garrett2024skillgen}; however, multiple robots expose opportunities for parallelism, which enable quicker task completion.
In the humanoid task in Figure~\ref{fig:methodology}, arm 1 and arm 2 can move simultaneously in order to pick their respective objects.
Still, in order to correctly complete many tasks, certain skills need to be performed in sequence.
In this task, arm 1 needs to pick the cup before it can pour.
The pour must occur before arm 2 grasps the bowl in order to ensure it receives the contents of the cup.
In other tasks, such as the coordinated lift in the Build Tower example in Figure~\ref{fig:overview} ({\it bottom left}), both arms must execute a synchronized skill on the same object.

To model these dynamics, we introduce temporal skill constraints, which specify partial orders between specific steps of two skills, generalizing past notions~\cite{jiang2024dexmimicgen,lin2026humanoidmimicgen} in a manner aligned with mathematical programming.
A temporal precedence constraint $p = \langle \psi, i, \psi', j \rangle$, where $0 \leq i < |\vec{s}_\psi|$ and $0 \leq j < |\vec{s}_{\psi'}|$, asserts that step $i$ of skill $\psi$ must execute no later than step $j$ of skill $\psi'$.
Let a skill plan $\langle \Psi, P \rangle$ be a set of robot skills $\psi \in \Psi$ and a set of temporal precedence constraints $p \in P$.
Temporal constraints can represent many different dynamics.
Consider the following examples\footnote{Let an index value of $-1$ be shorthand for the last step on a skill.} involving two skills $\psi, \psi'$.
We represent {\em sequential} relationships using:
$P = \{\langle \psi, -1, \psi', 0 \rangle\}.$
If the skills have no ordering, then they can be executed in {\em parallel}:
$P = \emptyset.$
A coordination relationship between the two skills can be expressed as:
\[P = \{\langle \psi, 0, \psi', 0 \rangle, \langle \psi', 0, \psi, 0 \rangle,
    \langle \psi, -1, \psi', -1 \rangle, \langle \psi', -1, \psi, -1 \rangle\},\]
where two inequalities at the start $i = j = 0$ and end $i = j = -1$ of the skills result in an equality constraint.

\section{ScheduleGen}

ScheduleGen solves a mixed discrete-continuous planning \& scheduling problem to generate demonstrations.
Figure~\ref{fig:methodology} outlines ScheduleGen's demonstration generation phases.
ScheduleGen takes as input the initial state $s_0$ and a skill plan $\langle \Psi, P \rangle$ (Section~\ref{sec:problem}).
Then, it operates in two phases.
First, it plans joint-space trajectories per robot by adapting each skill $\psi$ to the current state and connecting consecutive robot skills using
motion planning (Section~\ref{sec:planning}).
Then, it schedules execution of each action among the robots, satisfying temporal and collision constraints as well as minimizing cost functions like schedule duration (Section~\ref{sec:scheduling}).
Finally, we describe some GPU acceleration considerations in Section~\ref{sec:gpu}.

\begin{algorithm}[!htbp]
    \caption{ScheduleGen}
    \label{alg:schedulegen}
    \begin{algorithmic}[1]
    \begin{small}
        \Procedure{schedule-gen}{${\cal R}, p_0, S_*, {\cal D}; n$}
            \State $D \gets \emptyset$ \Comment{Initialize the successful demo dataset}
            \While{$|D| < n$} \Comment{Loop until the dataset is complete}
                \State $s_0 \sim p_0$; $d \sim {\cal D}$ \Comment{Sample the initial state and demo}
                \State $\langle \Psi, P \rangle \gets \Call{segment-skills}{d}$ \Comment{Segment the demo}
                \State ${\cal T} \gets \Call{serialized-planner}{{\cal R}, s_0, \langle \Psi, P \rangle}$
                \If{${\cal T} \neq \kw{None}$}
                    \State $\tau \gets \Call{schedule-robots}{{\cal R}, {\cal T}, P}$
                    \If{$\tau \neq \kw{None}$}
                        \State $\langle \vec{s}, \vec{a}, \vec{o} \rangle \gets \Call{execute}{s_0, \tau}$ \Comment{Execute the demo}
                        \If{$\vec{s}[-1] \in S_*$} \Comment{Check for demo success}
                            \State $D \gets D \cup \{\langle \vec{o}, \vec{a}  \rangle\}$ \Comment{Add the successful demo}
                        \EndIf
                    \EndIf
                \EndIf
            \EndWhile
            \State \Return $D$ \Comment{Return the generated demos}
        \EndProcedure
    \end{small}
    \end{algorithmic}
\end{algorithm}

Algorithm~\ref{alg:schedulegen} gives the high-level pseudocode for \proc{schedule-gen}.
It takes in the set of robots ${\cal R}$, initial state distribution $p_0$, goal state set $S_*$, set of source demonstrations ${\cal D}$, and a target number of demonstrations $n$.
Until $n$ successful demonstrations in $D$ are generated, it randomly samples an initial state $s_0$ and demonstration $d$.
The demonstration $d$ is segmented into a skill plan $\langle \Psi, P \rangle$ (done a priori in practice).
Then, \proc{schedule-gen} computes a collection of trajectories ${\cal T}$ for the robots ${\cal R}$ that independently perform each skill (Section~\ref{sec:planning}).
This collection is safely and correctly scheduled into a single multi-robot trajectory $\tau$ by \proc{schedule-robots}.
The trajectory is executed in simulation producing states $\vec{s}$, actions $\vec{a}$, and observations $\vec{o}$.
If terminal state $\vec{s}[-1]$ is in $S_*$, the demonstration is deemed a success and added to the successful demo dataset $D$.

\subsection{Skill Planning}\label{sec:planning}

ScheduleGen's first phase samples manipulation plans per robot that perform and connect the skills.
Algorithm~\ref{alg:serialized} presents the pseudocode for \proc{serialized-planner}, which computes a manipulation plan for each robot that performs its associated skills.
\proc{serialized-planner} takes as input the set of robots ${\cal R}$, initial state $s_0$, and skill plan $\langle \Psi, P \rangle$.
It returns a mapping ${\cal T}$ from each robot $r \in {\cal R}$ to a sequence of robot configuration states and controller targets.
First, \proc{serialized-planner} topologically sorts $\Psi$ into a sequence of skills by performing a Depth-First Search (DFS) using the sequential precedence constraints in $P$.
Then, it processes each skill $\psi = \langle r, f, \vec{s}_\psi, \vec{a}_\psi \rangle$ in resulting order.
Using the current pose of $s_0[f]$ for object frame $f$, it adapts the states $\vec{s}_\psi$ and actions $\vec{a}_\psi$ on the skill demonstration (Section~\ref{sec:skills}).
Each adapted end-effector pose $s'_\psi[e_r]$ serves as the target for iterative inverse kinematics, inducing a skill trajectory $\tau_\psi$.
Then, \proc{serialized-planner} plans a motion trajectory $\tau_m$ to the skill's initial robot configuration.
The motion and skill trajectories $\tau_m$ and $\tau_\psi$ are logged in ${\cal T}$ for robot $r$, and the robot and object states are updated in $s_0$.
Inverse kinematics and motion planning can condition their behavior on the current state $s_0$; however, collision constraints with other moving robots are deferred until Section~\ref{sec:scheduling}.
If inverse kinematics or motion planning fails, \proc{serialized-planner} returns \kw{None} and the candidate is rejected. Later attempts resample the initial state and source demonstration.

\begin{algorithm}[!htbp]
    \caption{Serialized Multi-Robot Planning}
    \label{alg:serialized}
    \begin{algorithmic}[1]
    \begin{small}
        \Procedure{serialized-planner}{${\cal R}, s_0, \langle \Psi, P\rangle$}
            \State ${\cal T} \gets \{r: [\;] \;\kw{for}\; r \in {\cal R} \}$ \Comment{Initial robot trajectories}
            \For{$\langle r, f, \vec{s}_\psi, \vec{a}_\psi \rangle \in \Call{topological-sort}{\Psi, P}$}
                \State $\tau_\psi \gets [\;]$ \Comment{Initialize skill trajectory}
                \For{$\langle s'_\psi, a'_\psi \rangle \in \proc{adapt-demo}(\vec{s}_\psi, \vec{a}_\psi, s_0[f])$}
                   \State $s'_\psi[r] \gets \proc{inverse-kinematics}(r, s'_\psi[e_r]; s_0)$
                   \State $a'_\psi[r] \gets \proc{inverse-kinematics}(r, a'_\psi[e_r]; s_0)$
                    \If{$s'_\psi[r] = \kw{None} \;\kw{or}\; a'_\psi[r] = \kw{None}$}
                        \State \Return $\kw{None}$
                    \EndIf
                    \State $\tau_\psi \gets \tau_\psi + [\langle s'_\psi, a'_\psi \rangle]$ \Comment{Add skill state and action}

                \EndFor
                \State $\langle s_\psi, \_ \rangle \gets \tau_\psi[0]$ \Comment{Initial skill state}
                \State $\tau_m \gets \proc{motion-planner}(r, s_0[r], s_\psi[r]; s_0)$
                \If{$\tau_m = \kw{None}$}
                    \State \Return $\kw{None}$
                \EndIf
                \State ${\cal T}[r] \gets {\cal T}[r] + \tau_m + \tau_\psi$ \Comment{Concatenate new trajectories}
                \State $s_0 \gets s'_\psi$; \Comment{Update the state}
            \EndFor
            \State \Return ${\cal T}$
        \EndProcedure
    \end{small}
    \end{algorithmic}
\end{algorithm}

\subsection{Skill Scheduling}\label{sec:scheduling}

The output of Section~\ref{sec:planning} is a collection of trajectories ${\cal T} = \{r: \tau_r \mid r \in {\cal R}\}$ for robots ${\cal R}$.
At this point, prior work would execute the trajectories sequentially~\cite{garrett2024skillgen} or simultaneously but myopically~\cite{lin2026humanoidmimicgen}, but these schemes result in inefficient or unsafe demonstrations because they don't reason about collision-free simultaneous motion.
For example, in Figure~\ref{fig:methodology}, even if arm 2 doesn't initiate its pick bowl skill, moving immediately to the bowl obstructs arm 1 from safely pouring the cup.
As a result, long-horizon scheduling is required to coordinate this behavior.

To produce high-quality multi-arm demonstrations, we propose {\em scheduling} the skills in order to minimize execution time while respecting task and collision constraints.
We accomplish this by implementing $\Call{schedule-robots}{{\cal R}, {\cal T}, P}$ using mathematical programming.
We define nonnegative time decision variables $\decision{t_r}$, where $|\decision{t_r}| = |{\cal T}[r]| = n_r$ for each robot $r$.
The objective is to minimize the {\em makespan} of the schedule $\decision{t_*}$, the maximum robot execution duration, subject to constraints:
$\decision{t_*} = \max_{r \in {\cal R}} \decision{t}_r[-1]$.
We formulate safe scheduling as a mathematical program.

The objective minimizes makespan together with the sum of robot end times, weighted by $w\geq0$.
The experiment in Table~\ref{tab:insert-tubes-scheduling-costs} ablates other secondary costs.
Constraint one sets the start time $\decision{t}_r[0]$ to be zero.
Constraint two ensures the makespan is no less than each end time $\decision{t}_r[-1]$.
Constraint three gives the minimum time step $\Delta t$ between waypoints.
The minimum time step prevents scheduling from increasing waypoint-to-waypoint speed; acceleration is not explicitly constrained by this formulation.
Constraint four enforces the temporal constraints $P$ between robots $r, r' \in {\cal R}$.
Constraint five is a disjunctive collision constraint: if configurations $q_r[i], q_{r'}[j]$ collide, robot $r$ must finish $i$ before robot $r'$ starts $j$ (plus a safety parameter $\epsilon > 0$) or vice versa.
Critically, the disjunctive constraint makes this a Mixed-Integer Linear Program (MILP) as there's a discrete decision for which order to impose whenever there's a collision.
We create binary decision variables indicating which clause is active and use the Big-M method to enforce that only one constraint is active. Appendix~\ref{app:collision-execution} gives the resulting inequalities.

\begin{small}
\begin{align}\label{eq:scheduling}
 \min_{\decision{t_*}} & \;\decision{t_*} + w\sum_{r \in {\cal R}} \decision{t}_r[-1] & \\
 \suchthat & 0 = \decision{t}_r[0] & \forall r \in {\cal R} \nonumber\\
 &  \decision{t}_r[-1] \leq \decision{t}_* & \forall r \in {\cal R}  \nonumber\\
 & \decision{t}_r[i] + \Delta t \leq \decision{t}_r[i+1], & \forall r \in {\cal R}, i \in [n_r-1] \nonumber\\
 & \decision{t}_{r}[i] \leq \decision{t}_{r'}[j], & \forall \langle \psi_r, i, \psi_{r'}', j \rangle \in P\nonumber\\
 & \decision{t}_{r}[i+1] + \epsilon \leq \decision{t}_{r'}[j] & \substack{\forall r \neq r' \in {\cal R},\ i \in [n_r-1],\\ j \in [n_{r'}-1]}\nonumber\\
 &\kw{ or }\; \decision{t}_{r'}[j+1] + \epsilon \leq \decision{t}_{r}[i], & \kw{if}\;\proc{collision}(q_r[i], q_{r'}[j])\nonumber
\end{align}
\end{small}

We extract a joint robot trajectory $\tau$ from the MILP by rounding the time variables $\decision{t}_r$, bucketing them in indices $\lfloor t / \Delta t \rfloor$, and maintaining the same set point per robot until the next waypoint.
If any robot start $q_r[0]$ or end $q_r[-1]$ states collide, then the MILP is infeasible.

\subsection{GPU Acceleration}\label{sec:gpu}

In order to formulate the MILP in section~\ref{sec:scheduling}, we need to be able to evaluate the collision constraints using \proc{collision}.
This requires checking all configuration pairs across robot trajectories, inducing $\mathcal{O}(|{\cal R}|^2 n_{\max}^2)$ collision checks for $n_{\max}=\max_r n_r$.
To combat this, we leverage GPU acceleration to evaluate these collisions in parallel, sometimes in a single batch call.
We frame robot-robot collision checking as a pairwise self-collision check in a composite robot made up of the two robots~\cite{garrett2025schedulestream}.
We use cuRobo's custom self-collision CUDA kernel~\cite{sundaralingam2023curobo,sundaralingam2026curobov2} for the sphere-sphere collision checks.
We use a tangent sphere inflation~\cite{inui2016shrinking} strategy to compute spherical decompositions of the robots.
We represent grasped objects during scheduling also as a set of spheres that temporarily augment the collision geometry of the robot.

\section{Experiments}
\label{sec:experiments}

We evaluate ScheduleGen on bimanual, four-arm, humanoid, benchmark-transfer, and real-world tasks.\footnote{Experiment videos are available on the \href{https://njawale42.github.io/ScheduleGen/}{ScheduleGen Website}.} Section~\ref{sec:experimental-setup} defines the common settings and task suite, Section~\ref{sec:experiments-sim} presents the capabilities enabled by ScheduleGen, Section~\ref{sec:schedulegen-analysis} analyzes its data-generation and policy behavior, and Section~\ref{sec:experiments-sim2real} reports real-world transfer.

\subsection{Experimental Setup}
\label{sec:experimental-setup}

\textbf{Data-Generation and Policy Training Settings.} Unless otherwise stated, we use five annotated source demonstrations per task, share them across variants and baselines, and collect 1,000 successful demonstrations per variant. We train flow-matching visuomotor~\cite{lipman2023flow} policies on the demonstrations and evaluate them on held-out resets. We report data-generation success, policy success, and mean steps to completion over successful rollouts. Appendix~\ref{app:policy-training} gives the policy training settings.

\textbf{Baselines.} We compare ScheduleGen with DexMimicGen~\cite{jiang2024dexmimicgen}, which transforms and composes per-arm demonstration segments independently, and DexMimicGen-Tuned, which adds manually specified task-level holds and ordering constraints
to reduce collisions caused by arm-arm overlap for the defined reset distributions.

\begin{figure}[p]
    \centering
    \includegraphics[width=\linewidth]{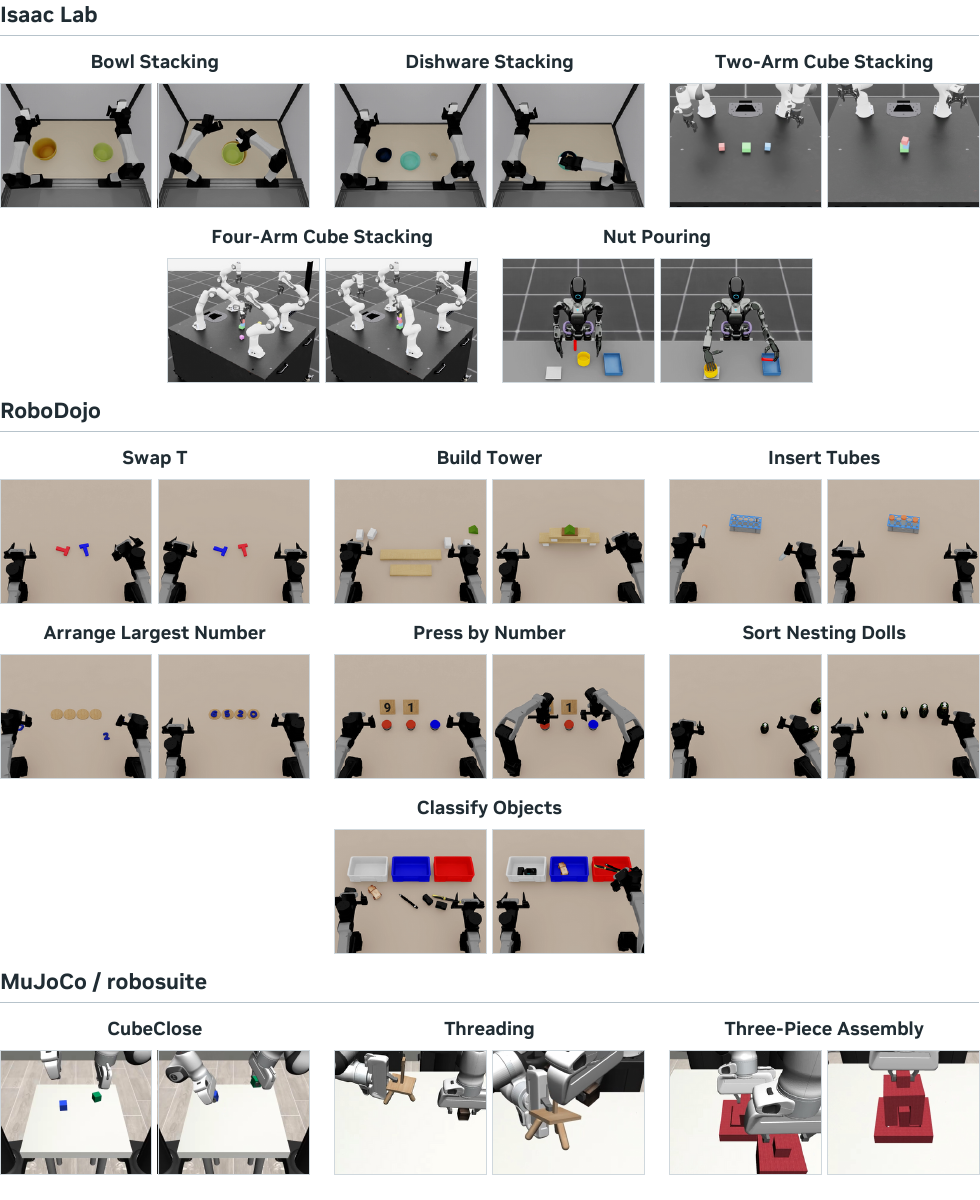}
    \caption{\textbf{Simulation task suite.} Five Isaac Lab, seven RoboDojo, and three robosuite simulated task families are shown. Paired views show task execution ({\it left}) and completion ({\it right}). CubeClose is evaluated under D0 and D1.}
    \label{fig:task-collage-all}
\end{figure}

\textbf{Tasks and Variants.} We define five task families in Isaac Lab~\cite{mittal2025isaac}: YAM Bowl and Dishware Stacking, DROID Two- and Four-Arm Cube Stacking, and fixed-waist GR1T2 Humanoid Nut Pouring. \emph{Normal} denotes the source layout, \emph{Cross} moves the arm-object assignments across the robot centerline while reusing the same source demonstrations, and \emph{Bin} adds a shallow bin as an obstacle/clutter condition on both Normal and Cross. Four-Arm Cube Stacking builds a five-object tower on a table or in a bin under a base reset support (Distribution I) and an expanded support (Distribution II).
We additionally evaluate four robosuite variants in MuJoCo and seven released RoboDojo tasks on the X5 bimanual setup~\cite{chen2026robodojo} (Fig.~\ref{fig:task-collage-all}). For RoboDojo policy evaluation and the aggregate, \emph{Nominal} uses held-out layouts from the training reset distribution, while \emph{Perturbed} uses the same normalized draws with object-pose perturbations doubled.

\begin{figure}[p]
    \centering
    \includegraphics[width=\linewidth]{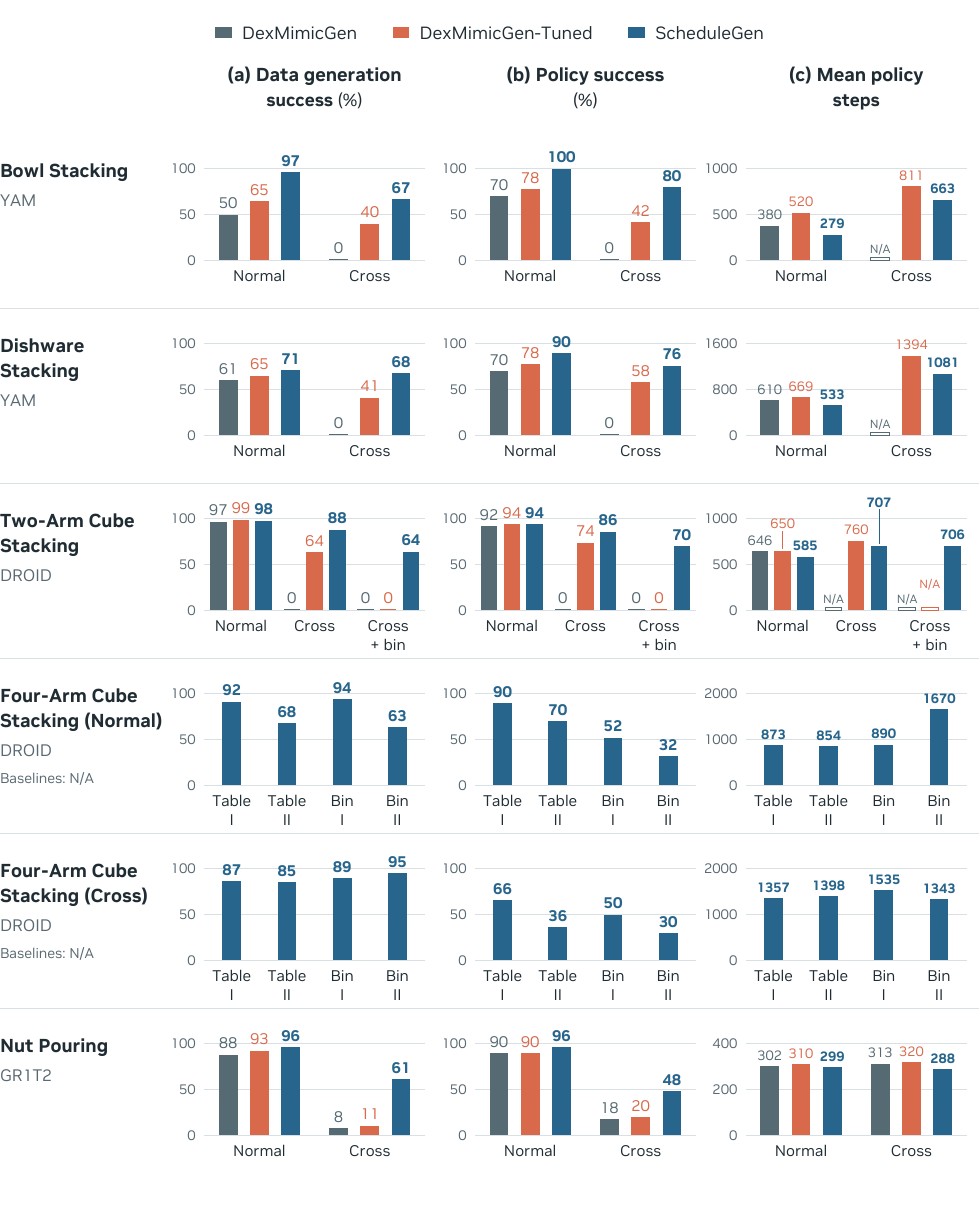}
    
    \caption{\textbf{Simulation data generation and policy evaluation.} Rows group tasks and embodiments; columns show data generation success, policy success, and mean steps over successful policy episodes. Methods share source demonstrations within each family. Policies use 50 evaluation episodes. Four-arm policies train separately for distributions I and II; DexMimicGen baselines are unsupported. Labels are rounded; bar heights retain the original values. Filled baseline markers denote zero; hollow markers denote unavailable steps. Unsupported four-arm baselines are omitted. Step axes use task-specific ranges.}
    \label{fig:sim-results}
    
\end{figure}

\subsection{ScheduleGen Features}
\label{sec:experiments-sim}

\textbf{ScheduleGen generates demonstrations across diverse tasks and robot embodiments.}
Across 17 Isaac Lab variants, ScheduleGen achieves 61.1-98.0\% data-generation success, exceeding 85\% on 10 variants. All policies trained on this data reach 30-100\% success (Fig.~\ref{fig:sim-results}). These results span bimanual YAM and DROID~\cite{khazatsky2024droid}, a four-arm DROID setup, and a GR1T2 humanoid. Data-generation success also remains high (75.5-100.0\%) across seven RoboDojo tasks on X5 (Fig.~\ref{fig:robodojo-generation-scheduling}).
On four robosuite variants in MuJoCo, BC-RNN-GMM policies reach 60-100\% success over 50 episodes from data generated at 24.7-84\% success (Table~\ref{tab:robosuite}).

\textbf{ScheduleGen generates collision-free demonstrations even in scenarios with inter-arm overlap, unlike prior work.}
The Cross-Arm variants introduce overlapping arm workspaces and broader object resets while reusing the Normal task demonstrations. DexMimicGen transfers each arm independently, while the manually selected holds in DexMimicGen-Tuned apply fixed timing across resets. ScheduleGen instead replans connecting motions and schedules collision-free waits for each reset. Across the four bimanual Cross and Cross-Bin variants, it reaches 64-88\% data-generation success and 70-86\% policy success. Two-Arm Cube Stacking uses a state-based BC-RNN, while Four-Arm Cube Stacking uses state-based flow-matching policies. DexMimicGen generates no valid trajectories on these variants, while DexMimicGen-Tuned remains below ScheduleGen throughout (Fig.~\ref{fig:sim-results}).

\textbf{ScheduleGen scales data generation and policy learning to multi-arm systems.} Four-Arm Cube Stacking uses motion-planned pick-and-place segments as source skills rather than teleoperation. ScheduleGen jointly schedules them across four arms, checking all six arm pairs while building a five-object tower on a table or in a narrow bin. We train a separate state-based flow-matching policy for each layout and reset distribution. Across eight settings, data-generation success is 63.4-95.1\% and policy success is 30-90\% (Fig.~\ref{fig:sim-results}). DexMimicGen is not applicable because its interface supports only bimanual generation.

\begin{table}[!htbp]
    \centering
    \caption{\textbf{Robosuite experiments.} ScheduleGen improves data generation success rate on three of four variants, improves or matches policy success wherever DexMimicGen (DexMG) trains a policy, and alone generates successful data on CubeClose D1.}
    \label{tab:robosuite}
    \normalsize
    \renewcommand{\arraystretch}{1.2}
    \begin{tabular}{llcc}
        \toprule
        \textbf{Task} & \textbf{Method} & \textbf{DGR (\%)} & \textbf{Policy SR (\%)} \\
        \midrule
        CubeClose D0
            & DexMG     & \textbf{87.6} & 98 \\
            & ScheduleGen & 84.0 & \textbf{100} \\
        \midrule
        CubeClose D1
            & DexMG     & 0.0 & --- \\
            & ScheduleGen & \textbf{72.3} & \textbf{94} \\
        \midrule
        Threading
            & DexMG     & 20.7 & \textbf{60} \\
            & ScheduleGen & \textbf{24.7} & \textbf{60} \\
        \midrule
        TPA
            & DexMG     & 14.0 & 66 \\
            & ScheduleGen & \textbf{39.7} & \textbf{80} \\
        \bottomrule
    \end{tabular}
\end{table}

\textbf{ScheduleGen uses scheduling costs to generate different coordination strategies.} On the RoboDojo Insert Tubes task, we compare Makespan with three secondary costs (Section~\ref{sec:scheduling}) under the same collision, precedence, synchronization, and per-arm ordering constraints. All four modes generate 500 successful demonstrations from the same five sources while maintaining 92.08-94.16\% data-generation success. Each dataset supports policy learning, with 40.0-50.0\% success over 60 held-out Nominal episodes per mode and 237.8-240.3 mean steps over successful rollouts (Table~\ref{tab:insert-tubes-scheduling-costs}).

\begin{table}[!htbp]
    \centering
    \caption{\textbf{Scheduling cost ablation.} Effect of scheduling costs on data generation and policy learning. Each cost generates 500 successful demonstrations. Policy success is measured over 60 Nominal episodes per cost; mean policy steps are calculated over successful rollouts. Appendix~\ref{app:costs} defines the secondary costs.}
    \label{tab:insert-tubes-scheduling-costs}
    \normalsize
    \setlength{\tabcolsep}{5pt}
    \renewcommand{\arraystretch}{1.2}
    \begin{tabular*}{\linewidth}{@{\extracolsep{\fill}}lrrr@{}}
        \toprule
        \textbf{Scheduling cost} &
        \shortstack[r]{\textbf{Data generation}\\\textbf{success (\%)}} &
        \shortstack[r]{\textbf{Policy}\\\textbf{success (\%)}} &
        \shortstack[r]{\textbf{Mean policy}\\\textbf{steps $\downarrow$}} \\
        \midrule
        \addlinespace[4pt]
        \parbox[c]{0.47\linewidth}{\textbf{Makespan}\\[-1pt]
            {\small Minimize just the schedule duration.}}
            & 94.16 & 43.3 & 239.8 \\
        \addlinespace[10pt]
        \parbox[c]{0.47\linewidth}{\textbf{Makespan + continuity}\\[-1pt]
            {\small Minimize the longest internal pause.}}
            & 93.46 & 43.3 & 240.2 \\
        \addlinespace[10pt]
        \parbox[c]{0.47\linewidth}{\textbf{Makespan + max-min velocity}\\[-1pt]
            {\small Maximize the slowest moving-edge velocity.}}
            & 92.08 & 40.0 & 237.8 \\
        \addlinespace[10pt]
        \parbox[c]{0.47\linewidth}{\textbf{Makespan + minimum active span}\\[-1pt]
            {\small Minimize summed arm-active spans.}}
            & 93.46 & 50.0 & 240.3 \\
        \addlinespace[4pt]
        \bottomrule
    \end{tabular*}
\end{table}

\textbf{ScheduleGen scales released demonstrations to externally authored tasks.} Agentic annotation converts five released RoboDojo demonstrations per task into seeds with per-arm contact boundaries, object-relative frames, and task constraints; ambiguous cases retain human review, and all annotations must pass replay and task-success checks. ScheduleGen then generates 500 successful training demonstrations per task at 75.5-100.0\% success (89.1\% mean). The resulting flow-matching policies complete all seven tasks under both conditions, averaging 47.9\% success on held-out Nominal layouts and 25.0\% under doubled pose perturbations over 20 episodes per task and condition (Fig.~\ref{fig:robodojo-generation-scheduling}).

\subsection{ScheduleGen Analysis}
\label{sec:schedulegen-analysis}

\par\smallskip
\textbf{ScheduleGen reduces the task-specific data collection and tuning needed for difficult layouts.}
We separately collect Cross-Arm demonstrations for YAM Bowl and Dishware Stacking. Each set takes the operator about 20 minutes, with additional work for annotation and hold tuning. To generate 1,000 demonstrations, DexMimicGen reaches 32.1\% and 11.2\% data-generation success and 30\% and 20\% policy success evaluated over 50 episodes. ScheduleGen instead reuses the five Normal demonstrations, reaching 66.7\% and 68.0\% data-generation success and 80\% and 76\% policy success. Task-matched teleoperation alone therefore does not close the gap: ScheduleGen retains the contact skills while replanning and scheduling free-space motion for each reset.

\par\smallskip
\Needspace{4\baselineskip}
\textbf{Policies trained on ScheduleGen data outperform the best DexMimicGen baseline across tasks and embodiments.}
Across nine variants supported by DexMimicGen, ScheduleGen policies improve success on eight and match the best baseline at 94\% on Normal Two-Arm Cube Stacking. Policy success rises from 42-78\% to 76-100\% on Bowl and Dishware Stacking, 0-74\% to 70-86\% on the Cross-Arm and Cross-Bin cube variants, and 20-90\% to 48-96\% on Humanoid Nut Pouring. ScheduleGen policies also use fewer steps whenever a baseline policy succeeds (Fig.~\ref{fig:sim-results}).

\par\smallskip
\textbf{ScheduleGen shortens RoboDojo trajectories by scheduling more arm motion in parallel.}
Across the 3,500 trajectories used for policy training, ScheduleGen reduces mean execution length by 2.3-11.1\% per task relative to the released RoboDojo trajectories, with a 7.5\% average reduction across tasks. Both arms are active together for 31.5\% of execution steps, compared with 15.5\% in the released demonstrations. To measure scheduling alone, we also apply the scheduler to all 700 released trajectories while keeping their paths and task order fixed. It finds valid schedules for 98.3\% of these trajectories and reduces total steps by 14.98\%
without replanning or physics replay. The mean reduction per episode is 16.44\% (Fig.~\ref{fig:robodojo-generation-scheduling}). Appendix~\ref{app:robodojo-source} gives the per-task measurements.

\begingroup

\begin{figure}[!htbp]
    
    \centering
    \includegraphics[width=\linewidth]{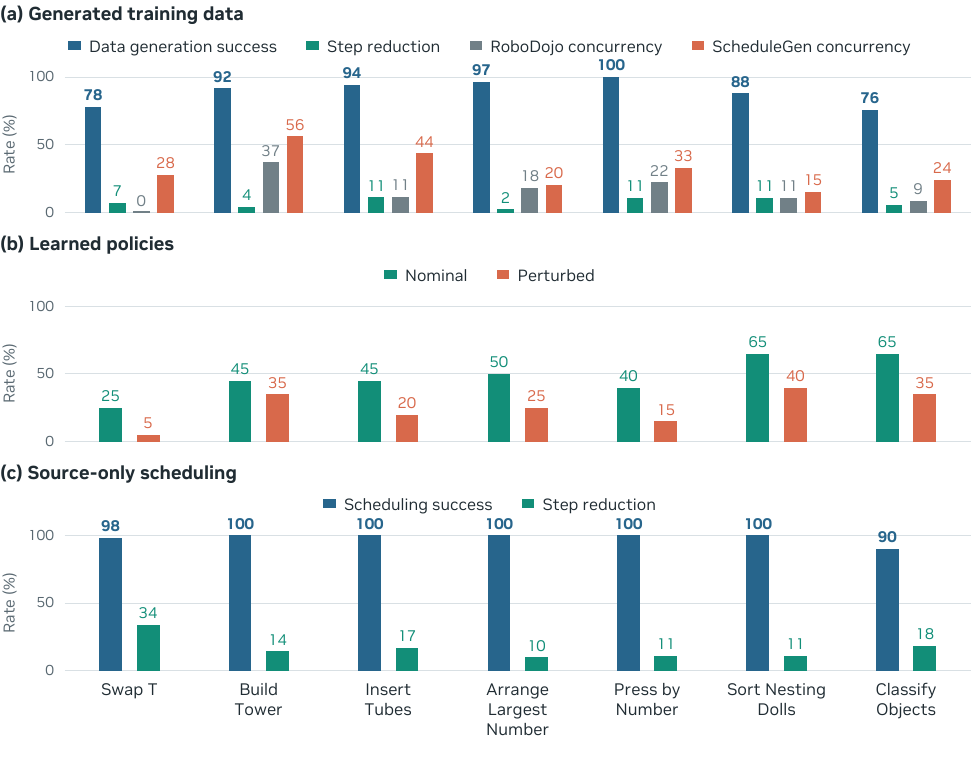}
    \caption{\textbf{RoboDojo results.} (a) Data generation success, reduction in mean steps relative to released demonstrations, and concurrent arm motion for 500 generated trajectories per task. (b) Policy success over 20 Nominal and 20 Perturbed episodes per task. (c) Scheduling success and mean per-trajectory step reduction for 700 fixed released trajectories (Table~\ref{tab:appendix-robodojo-source-scheduling}). Concurrency is the fraction of steps when both arms move; labels are rounded. A short baseline marker denotes zero.
    }
    \label{fig:robodojo-generation-scheduling}
    
\end{figure}
\endgroup

\FloatBarrier

\subsection{Real-World Evaluation}
\label{sec:experiments-sim2real}

We evaluate two transfer routes on the real YAM setup. In sim-to-real, policies are trained only on ScheduleGen simulation data. In real-to-sim-to-real, calibrated real demonstrations seed simulation data generation for Cube Stacking and Trash, and the resulting policies are returned to the same robot. All policies are trained in simulation without real-world fine-tuning. We use calibrated robot, camera, workspace, and object geometry, and randomize object pose, appearance, lighting, camera intrinsics and extrinsics, and background. Policies observe three RGB views and both arms' joint and gripper states, and predict joint-position and gripper targets.

\begingroup
\setlength{\intextsep}{8pt}

\begin{figure}[!htbp]
    \centering
    \includegraphics[width=\linewidth]{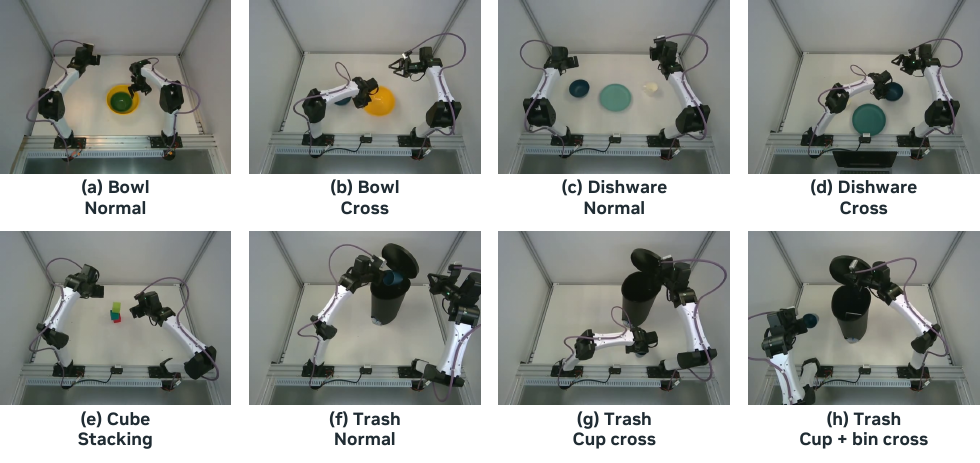}
    \caption{\textbf{ScheduleGen policies transfer to the real world.} Top: successful sim-to-real Bowl and Dishware Stacking rollouts (Normal and Cross). Bottom: real-to-sim-to-real Cube Stacking and Trash rollouts (Normal, Cup cross, and Cup + bin cross). Each policy is evaluated over 20 real-world episodes; Table~\ref{tab:real-world-transfer} reports success rates.
    }
    \label{fig:real-world-results}
\end{figure}
\endgroup

\begingroup
\setlength{\intextsep}{8pt}
\begin{table}[!htbp]
    \centering
\caption{\textbf{Simulation data-generation and real-world policy success.} Each policy is evaluated over 20 real-world episodes without real-world fine-tuning. Each variant uses 1,000 ScheduleGen-generated simulation demonstrations. Sim-to-real uses no real sources; real-to-sim-to-real uses calibrated real sources, collected only in the Normal layout for Trash. DGR: data generation success rate in simulation. SR: real-world policy success rate.
}
    \label{tab:real-world-transfer}
    \normalsize
    \setlength{\tabcolsep}{5pt}
    \renewcommand{\arraystretch}{1.2}
    \begin{tabular*}{\columnwidth}{@{\extracolsep{\fill}}llcc@{}}
        \toprule
        \textbf{Experiment} & \textbf{Task} & \shortstack{\textbf{Sim DGR}\\\textbf{(\%)} $\uparrow$} & \shortstack{\textbf{Real SR}\\\textbf{(\%)} $\uparrow$} \\
        \midrule
        Sim-to-real & Bowl Stacking (Normal) & 96.5 & 75 \\
                    & Bowl Stacking (Cross-Arm) & 66.7 & 80 \\
                    & Dishware Stacking (Normal) & 71.2 & 80 \\
                    & Dishware Stacking (Cross-Arm) & 68.0 & 75 \\
        \midrule
        Real-to-sim-to-real & Cube Stacking & 64.6 & 65 \\
                    & Trash (Normal) & 74.9 & 75 \\
                    & Trash (Cup cross) & 59.5 & 70 \\
                    & Trash (Cup + bin cross) & 63.6 & 50 \\
        \bottomrule
    \end{tabular*}
    
\end{table}
\endgroup

\Needspace{4\baselineskip}
\textbf{ScheduleGen's simulation data train policies that transfer directly to the real YAM.}
The sim-to-real policies use no real demonstrations. Over 20 episodes per task, they reach 75\% success on Normal Bowl Stacking, 80\% on Cross-Arm Bowl Stacking, 80\% on Normal Dishware Stacking, and 75\% on Cross-Arm Dishware Stacking (Fig.~\ref{fig:real-world-results}).

\Needspace{6\baselineskip}
\textbf{ScheduleGen expands a small set of real demonstrations into simulation datasets that train policies for real deployment.} After reconstructing and physics-validating five real Cube Stacking demonstrations, ScheduleGen generates a randomized simulation dataset whose policy reaches 65\% success over 20 real-world episodes.

For Trash, real source demonstrations collected only in the Normal layout seed all three variants. ScheduleGen generates 1,000 demonstrations per variant, with data generation success rates of 74.9\%, 59.5\%, and 63.6\% for Normal, Cup cross, and Cup + bin cross, respectively. The resulting policies reach 75\%, 70\%, and 50\% success over 20 real-world episodes per variant. Appendix~\ref{app:real-transfer} gives the metric definitions.

\FloatBarrier

\section{Conclusion}

We presented ScheduleGen, an automated demonstration generation system for multi-arm imitation learning. It combines object-centric skill adaptation, motion planning, and joint scheduling to preserve contact-rich behavior while satisfying task order, synchronization, and collision constraints. Experiments across 28 simulation variants show that ScheduleGen supports diverse embodiments and challenging layouts, and trains policies that outperform prior bimanual methods on most comparable tasks. The learned policies also transfer to eight real-world variants without real-world fine-tuning. These results show that adapting both motion and coordination can reduce task-specific data collection while supporting long-horizon, contact-rich multi-arm manipulation.

\section{Acknowledgments}
The authors would like to thank Desmond Kyu, Justin Tran, and Ramanpreet Singh for data collection and model evaluation; Matin Nikoui and Lion Park for hardware support; Tianyuan Dai for help with simulation and domain randomization infrastructure; and Tuur Stuyck, Peter Du, Yashraj Narang, and Fabio Ramos for feedback and valuable discussions.
\FloatBarrier
\begingroup
\small
\setlength{\bibsep}{3pt plus 1pt}
\bibliographystyle{plainnat}
\bibliography{references}
\endgroup
\clearpage
\appendix
\section*{Appendix}
\addcontentsline{toc}{section}{Appendix}
\noindent
\hyperref[app:algorithm]{A. Planning and Scheduling Details}\\
\hyperref[app:costs]{B. Scheduling Costs}\\
\hyperref[app:setup]{C. Additional Results and Evaluation}\\
\hyperref[app:policy-training]{D. Policy Training}
\medskip
\section{Planning and Scheduling Details}\label{app:algorithm}
Algorithms~\ref{alg:schedulegen} and~\ref{alg:serialized} give the generation loop and skill planner. Here we describe how states and actions form the arm paths and expand the collision constraints in Section~\ref{sec:scheduling}.

\subsection{Path Construction}\label{app:path-construction}
Algorithm~\ref{alg:serialized} processes skills in a topological order of their sequential precedence constraints, updating object poses for later skills. This order is used to construct the planning scene; the arms can still execute in parallel after scheduling.

The adapted state sequence defines the configurations used for planning and collision checking, while the adapted action sequence supplies the controller targets for replay. We solve inverse kinematics for both sequences and retain the demonstrated finger states and actions. Connecting motions are inserted between skills as in Fig.~\ref{fig:methodology}(a). We then map each skill step to its index in the full arm path so that the scheduler can enforce the original temporal constraints.

\subsection{Collision Constraints and Execution}\label{app:collision-execution}
Let $C$ be the set of waypoint pairs $(r,i,r',j)$ for which $\proc{collision}(q_r[i],q_{r'}[j])$ holds in Eq.~\ref{eq:scheduling}. For each pair, a binary variable $b_{r,i,r',j}$ selects which arm moves through the conflict first. The Big-M constraints are
\begin{equation}
\begin{aligned}
 \decision{t}_r[i+1]+\epsilon &\leq \decision{t}_{r'}[j]+M b_{r,i,r',j}, \\
 \decision{t}_{r'}[j+1]+\epsilon &\leq \decision{t}_r[i]+M(1-b_{r,i,r',j}).
\end{aligned}
\label{eq:appendix-collision-order}
\end{equation}
If $b_{r,i,r',j}=0$, the first inequality is enforced; if it is 1, the second is enforced. $M$ must be large enough to make the other inequality inactive over the scheduling horizon. These constraints preserve waypoint order and the task's precedence and synchronization constraints while changing path timing. They apply to both connecting motions and contact skills, including the approach conflict in the Nut Pouring example (Fig.~\ref{fig:methodology}(b--c)). A feasible schedule is replayed in simulation to check contact behavior and task completion before the demonstration is accepted.

\section{Scheduling Costs}\label{app:costs}
Equation~\ref{eq:scheduling} minimizes makespan together with the weighted sum of arm end times. This section defines the secondary costs compared in Table~\ref{tab:insert-tubes-scheduling-costs}. They change path timing while preserving the task, collision, and per-arm ordering constraints.

\subsection{Primary and Secondary Objectives}
Let $\mathbf t$ collect the waypoint times $\decision{t}_r[i]$, and let $\mathcal S_{\mathrm{feas}}$ be the schedules satisfying the constraints in Eq.~\ref{eq:scheduling}. As in the main formulation, makespan is the latest arm end time, $\max_{r\in\mathcal R}\decision{t}_r[n_r-1]$.

For the cost ablation, we minimize makespan first and then optimize a secondary cost among schedules with that same makespan:
\begin{equation}
\begin{aligned}
 \decision{t}_*^{\mathrm{min}} &= \min_{\mathbf t\in\mathcal S_{\mathrm{feas}}}
       \max_{r\in\mathcal R} \decision{t}_r[n_r-1], \\
 \mathbf t_c^* &\in \arg\min_{\substack{\mathbf t\in\mathcal S_{\mathrm{feas}}\\
       \max_{r\in\mathcal R}\decision{t}_r[n_r-1]=\decision{t}_*^{\mathrm{min}}}} c(\mathbf t).
\end{aligned}
\label{eq:appendix-cost-order}
\end{equation}
Makespan alone uses the first stage. The other modes also use the second stage, so their secondary objective cannot increase the minimum makespan. This differs from the weighted objective in Eq.~\ref{eq:scheduling}, which balances makespan against arm end times. Schedules with the same makespan can still differ in when each arm moves, waits, and finishes; the secondary costs distinguish these timings.

\subsection{Secondary Cost Definitions}
For consecutive waypoints, define the interval in control ticks and the joint-space distance as
\[
 g_r[i]=\frac{\decision{t}_r[i+1]-\decision{t}_r[i]}{\Delta t}\geq1,
 \qquad
 \ell_r[i]=\lVert q_r[i+1]-q_r[i]\rVert_2.
\]
The lower bound on $g_r[i]$ follows from the minimum time step in Eq.~\ref{eq:scheduling}.

\begin{subequations}\label{eq:appendix-secondary-costs}
\paragraph{Continuity.}
The set $\mathcal E_{\mathrm{int}}$ contains edges within an arm's active interval, excluding its launch delay and terminal hold. The continuity cost is
\begin{equation}
 c_{\mathrm{cont}}(\mathbf t)
   =\max_{(r,i)\in\mathcal E_{\mathrm{int}}}\bigl(g_r[i]-1\bigr).
\label{eq:appendix-continuity-cost}
\end{equation}
The quantity $g_r[i]-1$ measures the interval beyond an edge's minimum one-tick duration. Minimizing its largest value discourages long internal pauses. Excluding launch delays and terminal holds allows an arm to wait before starting or after finishing. In Nut Pouring (Fig.~\ref{fig:methodology}), the bowl arm can delay its approach until the other arm finishes pouring, then move without a long internal pause.

\paragraph{Max-min velocity.}
The set $\mathcal E_{\mathrm{move}}$ contains edges with $\ell_r[i]>\delta_q$, where $\delta_q$ is a joint-motion threshold, separate from the collision time margin $\epsilon$. The cost is
\begin{equation}
 c_{\mathrm{vel}}(\mathbf t)
   =\max_{(r,i)\in\mathcal E_{\mathrm{move}}}\frac{g_r[i]}{\ell_r[i]}.
\label{eq:appendix-velocity-cost}
\end{equation}
For a moving edge, joint-space speed is $\ell_r[i]/(g_r[i]\Delta t)$. Since $\Delta t$ is fixed, minimizing the largest $g_r[i]/\ell_r[i]$ maximizes the slowest moving-edge speed. Stationary edges are excluded to avoid division by zero, so this cost does not directly penalize their duration. Faster moving edges can therefore coexist with long pauses (Appendix~\ref{app:cost-example}).

\paragraph{Minimum active span.}
Let $u_r^{\mathrm{first}}$ and $u_r^{\mathrm{last}}$ be the first and last control ticks that advance a moving edge. The cost is
\begin{equation}
 c_{\mathrm{span}}(\mathbf t)
   =\sum_{r\in\mathcal R}\bigl(u_r^{\mathrm{last}}-u_r^{\mathrm{first}}+1\bigr).
\label{eq:appendix-active-span-cost}
\end{equation}
The span $u_r^{\mathrm{last}}-u_r^{\mathrm{first}}+1$ counts control ticks from an arm's first movement through its last. Minimizing the sum favors shorter active intervals and discourages waits between movements. Launch delays and terminal holds do not contribute to this cost. Continuity targets the longest internal pause; minimum active span also accounts for the accumulated duration of several shorter pauses.
\end{subequations}

\Needspace{20\baselineskip}
\section{Additional Results and Evaluation}\label{app:setup}
This section expands the experiments in Section~\ref{sec:experiments}. We examine source demonstration reuse, explain how released demonstrations become skill seeds, and show how scheduling changes coordination and execution length.

\subsection{Evaluation Metrics}\label{app:robosuite}\label{app:real-transfer}
Data generation success rate is the fraction of attempts that produce a successful demonstration after simulation replay. Policy success rate is measured over held-out evaluation episodes; mean policy completion steps use successful rollouts only. In real-world transfer, generation is evaluated in simulation and policy success is evaluated on the robot. Aggregate simulation rates average the 28 variants equally and use Nominal evaluation for RoboDojo.

In Table~\ref{tab:robosuite}, the CubeClose D1 dash means that no DexMimicGen policy was trained because generation produced no successful demonstrations. On CubeClose D0, ScheduleGen has lower data generation success but higher policy success. The two rates measure different stages and need not rank methods identically.

\subsection{Source Demonstration Reuse}\label{app:source-reuse}
The Cross-Arm comparison in Section~\ref{sec:schedulegen-analysis} tests whether collecting demonstrations directly in the target layout closes the gap to ScheduleGen. DexMimicGen uses newly collected Cross-Arm demonstrations for Bowl and Dishware Stacking, while ScheduleGen reuses the five Normal demonstrations. Both methods are evaluated on the Cross-Arm variants, where the arms reach across the robot centerline.

\begin{figure}[H]
\centering
\includegraphics[width=\linewidth]{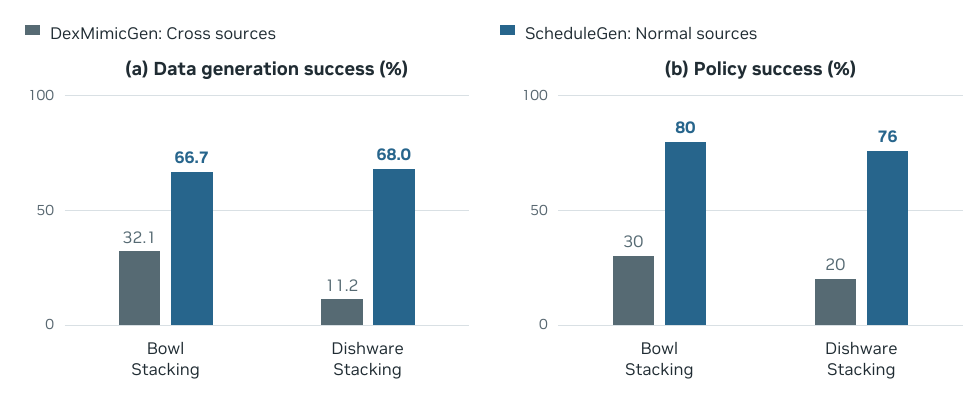}
\caption{\textbf{Source layout and Cross-Arm performance.} DexMimicGen receives newly collected Cross-Arm sources; ScheduleGen reuses Normal sources. The comparison reports data generation success and learned-policy success on the same two task variants. Each policy is evaluated over 50 episodes.}
\label{fig:appendix-source-reuse}
\end{figure}

Figure~\ref{fig:appendix-source-reuse} shows that task-matched sources alone do not recover the performance of ScheduleGen. A source demonstration provides the contact skills, but its connecting motions and timing may no longer be suitable when objects move. ScheduleGen plans those motions and coordinates the arms for each sampled scene, reducing the need to recollect demonstrations for a new layout.

\Needspace{10\baselineskip}
\subsection{RoboDojo Demonstration Transfer}\label{app:demonstration-transfer}
RoboDojo demonstrations provide RGB observations and joint actions. We use calibrated object geometry to reconstruct the scene in simulation and check the transferred motion through physics replay. Skill annotations specify per-arm contact boundaries, object-relative frames, and temporal constraints. A seed is used for generation only after replay and task-success checks.

\begingroup
\setlength{\intextsep}{6pt}
\begin{figure}[H]
\centering
\includegraphics[width=\linewidth]{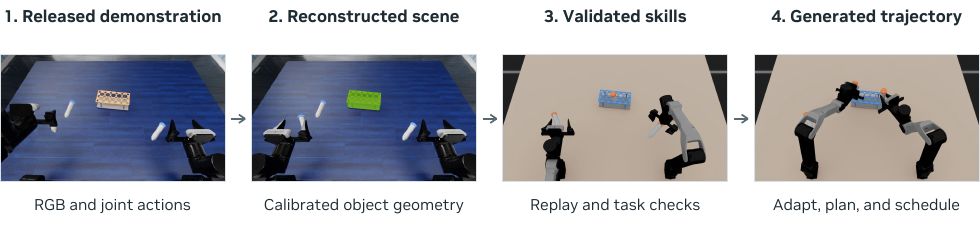}
\caption{\textbf{RoboDojo demonstration transfer.} Insert Tubes illustrates scene calibration, skill validation through replay, and trajectory generation. The same procedure supplies the five source demonstrations used for each RoboDojo task.}
\label{fig:appendix-demonstration-transfer}
\end{figure}
\endgroup

Algorithms~\ref{alg:schedulegen} and~\ref{alg:serialized} reuse these contact skills and task constraints in new scenes. ScheduleGen adapts the skills, plans connecting motions, and schedules the arms; generated demonstrations must pass replay before entering the training set.

\Needspace{10\baselineskip}
\subsection{Effects of Scheduling Costs}\label{app:cost-example}
Figure~\ref{fig:appendix-cost-example} compares scheduling costs on the same Insert Tubes skills and target scene. All modes finish their paths in 10.44~s (261 control ticks at 25~Hz); a shared 25-tick final hold brings each successful replay to 11.44~s.

An arm's active span runs from its first movement through its last, including internal pauses; panel (b) sums the spans over both arms. Continuity and minimum active span remove internal pauses and shorten this sum from 17.00 to 15.40~s. Max-min velocity permits a longer pause, showing that improving moving-edge speed need not give continuous motion. Appendix~\ref{app:costs} defines the costs; Table~\ref{tab:insert-tubes-scheduling-costs} reports policy results.

\begingroup
\setlength{\intextsep}{6pt}
\begin{figure}[H]
\centering
\includegraphics[width=0.90\linewidth]{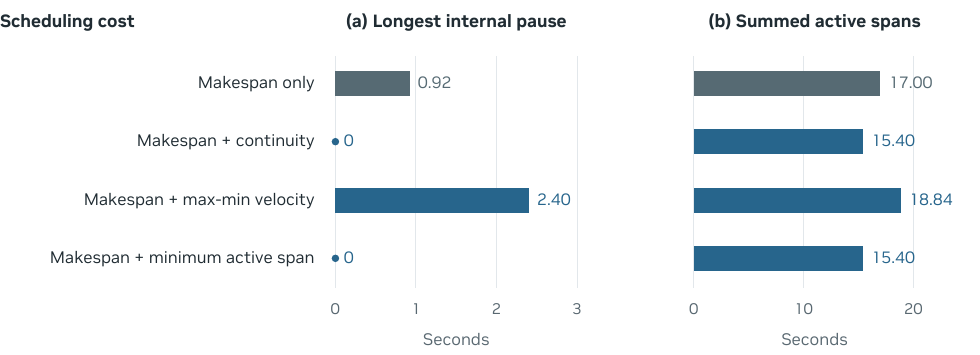}
\caption{\textbf{Scheduling costs change motion timing at the same completion time.} Both measures use seconds; lower is better. Summed spans can exceed completion time because arms move in parallel. Pauses and spans use a $10^{-3}$-rad joint-motion threshold.}
\label{fig:appendix-cost-example}
\end{figure}
\endgroup

\Needspace{600pt}
\subsection{Retiming Released RoboDojo Demonstrations}\label{app:robodojo-source}
We use the released RoboDojo demonstrations to ask whether ScheduleGen can improve coordination when the arm paths are already available. To isolate scheduling, we keep the per-arm paths and task order fixed for 100 demonstrations per task and optimize when the arms advance along those paths. Independent motions can overlap when task and collision constraints allow it; dependent motions must still wait. No new geometric paths are planned.

The scheduler finds feasible timings for 688 of the 700 trajectories (98.3\%). Here, feasibility means satisfying Eq.~\ref{eq:scheduling}; task completion is not checked through physics replay. Table~\ref{tab:appendix-robodojo-source-scheduling} compares each successfully scheduled trajectory with its original released version, so both step counts refer to the same underlying paths.

\begin{table}[H]
\centering
\caption{\textbf{Retiming released RoboDojo demonstrations.} Mean step counts use the source and scheduled versions of each successfully scheduled trajectory. Paths and task order are fixed. The last column gives the mean percentage reduction per trajectory.}
\label{tab:appendix-robodojo-source-scheduling}
\normalsize
\renewcommand{\arraystretch}{1.2}
\setlength{\tabcolsep}{3pt}
\begin{tabular*}{\linewidth}{@{\extracolsep{\fill}}lrrrr@{}}
\toprule
\textbf{Task} & \shortstack{\textbf{Scheduled/}\\\textbf{attempted}} & \shortstack{\textbf{Source}\\\textbf{steps}} & \shortstack{\textbf{Scheduled}\\\textbf{steps}} & \shortstack{\textbf{Mean reduction}\\\textbf{(\%)}}\\
\midrule
Swap T & 98/100 & 276.88 & 183.55 & 33.71\\
Build Tower & 100/100 & 726.78 & 622.51 & 14.34\\
Insert Tubes & 100/100 & 322.39 & 267.66 & 16.99\\
Arrange Largest Number & 100/100 & 583.33 & 523.59 & 9.92\\
Press by Number & 100/100 & 389.21 & 346.87 & 11.14\\
Sort Nesting Dolls & 100/100 & 590.13 & 522.82 & 11.20\\
Classify Objects & 90/100 & 689.62 & 572.87 & 18.31\\
\midrule
All tasks & 688/700 & --- & --- & 16.44\\
\bottomrule
\end{tabular*}
\end{table}
Retiming reduces mean step counts for every task. The released timings can already include concurrent arm motion, so these reductions show improvements over the source coordination rather than gains relative to fully serial execution.

The scheduled versions use 14.98\% fewer steps in total, with a mean reduction per trajectory of 16.44\%. These summarize the same paired comparisons in different ways. Let $\mathcal J$ contain the successfully scheduled trajectories, with source and scheduled step counts $L_k$ and $\widetilde L_k$ for $k\in\mathcal J$:
\begin{equation}
\begin{aligned}
 R_{\mathrm{pooled}} &= 100\left(1-
       \frac{\sum_{k\in\mathcal J}\widetilde L_k}
            {\sum_{k\in\mathcal J}L_k}\right), \\
 R_{\mathrm{episode}} &= \frac{100}{|\mathcal J|}
       \sum_{k\in\mathcal J}\left(1-\frac{\widetilde L_k}{L_k}\right).
\end{aligned}
\label{eq:appendix-step-reductions}
\end{equation}
The pooled reduction weights trajectories by source length, so longer demonstrations contribute more to the result. The mean per-trajectory reduction gives each demonstration equal weight, regardless of its original length. Both use the same 688 trajectories. Failed schedules count toward the scheduling success rate but are excluded from length comparisons.

In Fig.~\ref{fig:robodojo-generation-scheduling}(a), ScheduleGen adapts source skills to sampled scenes and checks generated demonstrations through replay. Here, each released trajectory is compared with a retimed version of itself. Together, the experiments show how scheduling contributes within the generation pipeline and when applied directly to existing arm paths.

\Needspace{42\baselineskip}
\section{Policy Training}\label{app:policy-training}
\paragraph{YAM visuomotor policies.}
YAM policies use three RGB views (left wrist, top, and right wrist) and a 14-dimensional absolute joint and gripper state to predict 14-dimensional absolute joint and gripper targets. Table~\ref{tab:appendix-flow-matching} lists the training settings. Scene randomization for real-world transfer is described in Section~\ref{sec:experiments-sim2real}.

\begin{table}[H]
    \centering
    \caption{\textbf{YAM visuomotor policy hyperparameters.}}
    \label{tab:appendix-flow-matching}
    \normalsize
    \renewcommand{\arraystretch}{1.2}
    \begin{tabular}{@{}p{0.34\columnwidth}p{0.60\columnwidth}@{}}
        \toprule
        \textbf{Component} & \textbf{Setting} \\
        \midrule
        Image encoder & DINOv3 ViT-L/16, fine-tuned \\
        Inputs & Three RGB views and 14-D state \\
        Output & 14-D absolute action, horizon 50 \\
        DiT & 32 layers; 16 heads; head dim. 64; hidden dim. 1024; dropout 0.2 \\
        Flow noise & $\mathrm{Beta}(1.5,1.0)$; $s=0.999$; 1,000 time buckets \\
        Inference & Four flow integration steps \\
        Optimizer & Fused AdamW; learning rate $10^{-4}$; weight decay $10^{-5}$; max grad. norm 1 \\
        Schedule & Cosine decay; 5\% warmup; 50,000 updates \\
        Batch & 32 per GPU; 8 GPUs; accumulation 1 (global 256) \\
        Precision & bfloat16, TF32, gradient checkpointing \\
        Checkpoints & Every 1,000 updates \\
        Trainable modules & Vision tower, DiT, and state/action adapters; language self-attention disabled \\
        Image augmentation & Resize shortest edge to 224; random 0.95 crop; brightness 0.3, contrast 0.4, saturation 0.5, hue 0.12 \\
        Seed & 42 \\
        \bottomrule
    \end{tabular}
\end{table}

\paragraph{BC policies.}
State-based BC uses a two-layer LSTM (hidden size 1,000, sequence length 10) with a five-mode Gaussian-mixture (GMM) or deterministic MSE head. Both use Adam, batch size 100, and a nominal 1,500-epoch budget. Initial learning rates are $5\times10^{-4}$ for GMM and $10^{-3}$ for MSE, halved at epochs 800 and 1,200.

Image BC uses the same recurrent core with one ResNet-18 and SpatialSoftmax encoder per view, a batch size of 16, and a nominal 600-epoch training budget. GMM policies use learning rate $10^{-4}$; MSE policies use $2\times10^{-4}$. Learning rates are halved at epochs 400 and 500. Task-specific checkpoints are selected using held-out rollout evaluation.

\end{document}